\documentclass{ecai}  % use option [doubleblind] for double blind submission and hiding the authors section

\usepackage{graphicx}
\usepackage{latexsym}
\usepackage{booktabs}
\usepackage[normalem]{ulem}
\useunder{\uline}{\ul}{}
\usepackage{array}
\newcolumntype{P}[1]{>{\centering\arraybackslash}p{#1}}
\newcommand{\specialcell}[2][c]{%
  \begin{tabular}[#1]{@{}P{1.2cm}@{}}#2\end{tabular}}

\newcommand{\specialcelltwo}[2][c]{%
  \begin{tabular}[#1]{@{}P{1.8cm}@{}}#2\end{tabular}}

\newcommand{\ourmodel}[1]{UCCIX}
\newcommand{\pretrainsubset}[1]{PT}

\ecaisubmission      % inserts page numbers. Use only for submission of paper.
\begin{document}

\begin{frontmatter}

\title{\ourmodel{}: Irish-eXcellence Large Language Model}

\author[A]{\fnms{Khanh-Tung}~\snm{Tran}}
\author[A]{\fnms{Barry}~\snm{O'Sullivan}\thanks{Corresponding Author. Email: b.osullivan@cs.ucc.ie}}
\author[A]{\fnms{Hoang D.}~\snm{Nguyen}} % use of \orcid{} is optional

\address[A]{University College Cork, Cork, Ireland}

\begin{abstract}
    The development of Large Language Models (LLMs) has predominantly focused on high-resource languages, leaving extremely low-resource languages like Irish with limited representation. This work presents \ourmodel{}, a pioneering effort on the development of an open-source Irish-based LLM. We propose a novel framework for continued pre-training of LLMs specifically adapted for extremely low-resource languages, requiring only a fraction of the textual data typically needed for training LLMs according to scaling laws. Our model, based on Llama 2-13B~\cite{abs-2307-09288}, outperforms much larger models on Irish language tasks with up to 12\% performance improvement, showcasing the effectiveness and efficiency of our approach. We also contribute comprehensive Irish benchmarking datasets, including IrishQA, a question-answering dataset, and Irish version of MT-bench~\cite{zheng2023judging}. These datasets enable rigorous evaluation and facilitate future research in Irish LLM systems. Our work aims to preserve and promote the Irish language, knowledge, and culture of Ireland in the digital era while providing a framework for adapting LLMs to other indigenous languages.
\end{abstract}
    
\end{frontmatter}
\setcounter{footnote}{-1}

\section{Introduction and Contribution}
    Large Language Models (LLMs) have demonstrated exceptional performance in a wide range of natural language processing (NLP) tasks, capable of understanding and generating human-like text. Prominent models, whether closed-source like ChatGPT, Google Gemini, or open-source variants such as Llama 2~\cite{abs-2307-09288} and BLOOM~\cite{bloom}, predominantly cater to popular languages like English, however, are limited in supporting for other indigenous languages. Moreover, the training of such LLMs often demands vast amounts of text data, numbering in the hundreds of billions or even trillions of tokens, adhering to recent scaling laws \cite{hoffmann2022an,muennighoff2023scaling}. This poses a significant challenge for most languages, particularly those with extremely limited resources. For instance, consider the case of Irish, even though it is the first official language of the Republic of Ireland, it is extremely underpresented, in both daily usages and on the Internet space. Classified as Definitely Endangered by UNESCO \cite{Unesco2010-st}, Irish faces an uphill battle for preservation and digitization. 

\begin{figure}[!t]
    \centerline{\includegraphics[width=1.0\linewidth]{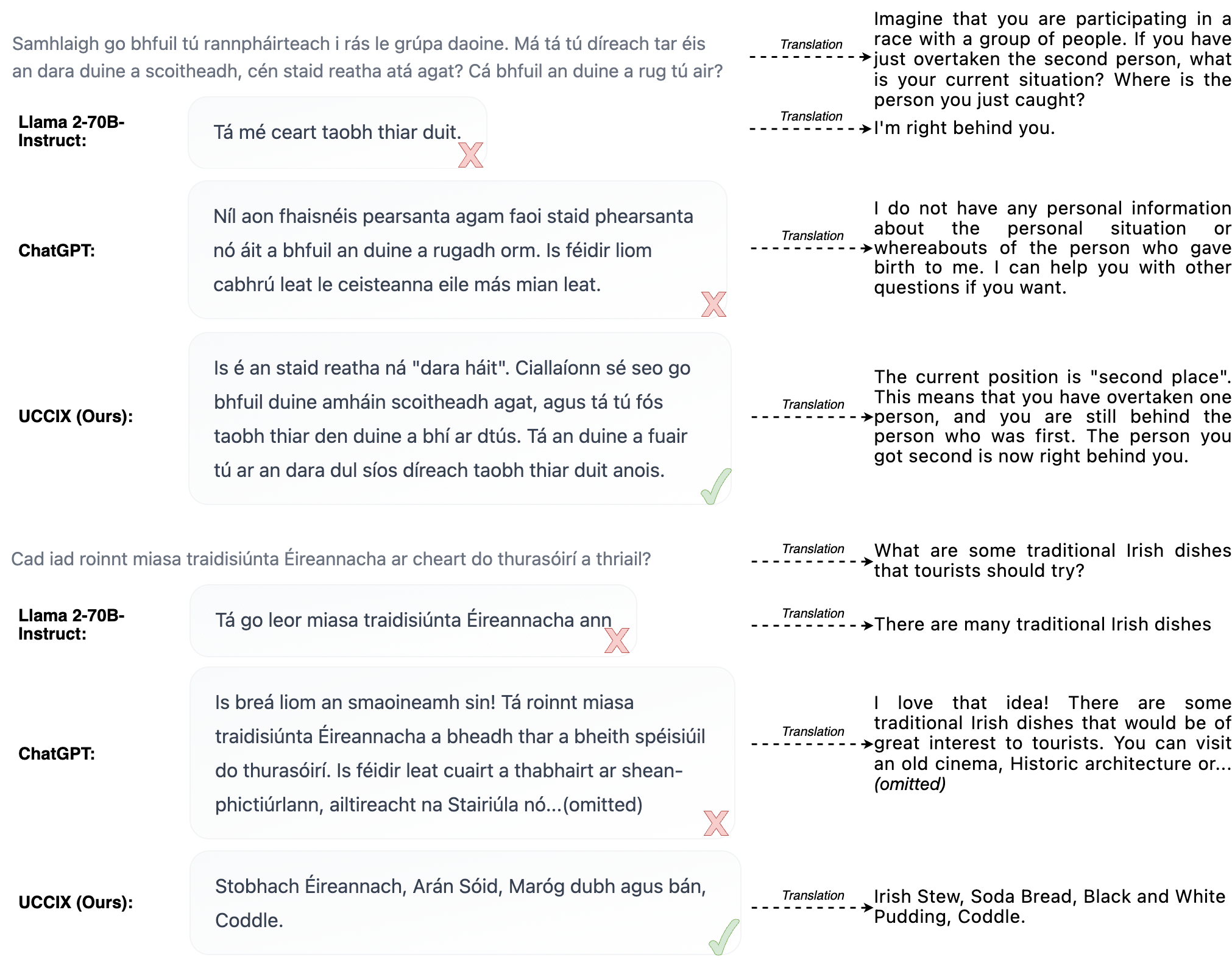}}
    \caption{Example of responses generated by our model, \ourmodel{}, and other baselines\protect\footnotemark. We demonstrate the English translation version on the right.}
    \label{fig:example}
    \end{figure}

    In this work, we attempt to develop open-source Irish-based LLMs with up to tens of billions of parameters for a language with only hundreds of million tokens of text \cite{gaBert}, only a fraction of the data typically required for training. This challenge presents an extreme scenario that has been overlooked in relevant works. For instance, FinGPT \cite{luukkonen-etal-2023-fingpt} utilized 300B Finnish tokens in order to be able to continued pre-train the base BLOOM model, and \cite{xu2024a} employed up to 20B tokens for five non-English languages. Similarly, Llemma \cite{azerbayev2024llemma} leveraged up to 200B tokens in mathematical topics for adapting CodeLlama. Approaches suggesting synthetic data generation, such as \cite{gunasekar2023textbooks,maini2024rephrasing}, are resource-intensive, and the quality of off-the-shelf LLMs in generating Irish text remains uncertain. BLOOM+1 \cite{yong-etal-2023-bloom} explored language adaptation at smaller model scales, resulting in modest performance gains. gaBERT \cite{gaBert}, a mono-Irish language model, operates at a much smaller scale with less than 300 million parameters. To address these limitations, we present \emph{\ourmodel{}}, the first Irish-based LLMs, trained on our curated set of Irish training datasets, covering broad domains and topics. We carry out experiments with the language adaptation strategy of continued pre-training, including tailored data mixture and scheduler during continued pre-training. 
    An example of the interactions with our LLMs is provided in Figure~\ref{fig:example}.
    We hope \ourmodel{} can be used as a tool to preserve the language, knowledge, and culture of Irish in the digital era. Moreover, while Irish is our case study in this research, we believe our framework can be adapted as a blueprint for language adaptation of large models to other low-resource languages.

    \footnotetext{Demo webpage: \url{https://aine.chat/}}

    Developing an LLM for an extremely low-resource language presents challenges not only in terms of training resources but also in the evaluation process. To our best knowledge, there is a scarcity of benchmarking datasets specifically tailored for evaluating LLMs in the context of low-resource languages \cite{10.1145/3641289}. This limitation poses a significant challenge in effectively assessing the performance and capabilities of generative language models designed for languages with limited textual data, such as Irish. Existing benchmark datasets that support Irish are often focused on traditional NLP tasks \cite{lankford-etal-2021-machine,gaBert,adelani-etal-2024-sib} and may not fully capture the capabilities of modern LLMs, such as language understanding and reasoning. This poses a need for tailoring a new set of benchmark data for Irish-based LLMs. To address this gap, we create new datasets that enable comprehensive benchmarking across various scenarios and settings, such as open-book question answering, and multi-turn conversational interactions.

    Our contributions:
    \begin{itemize}
        \item An innovative framework for continued pre-training of large language models for extremely low-resource languages.
        \item \ourmodel{}, an Irish-based LLM, based on Llama 2-13B. \ourmodel{} outperforms much larger models on Irish tasks with up to 12\% performance improvement, demonstrating the effectiveness and efficiency of our proposed framework.
        \item Irish benchmarking datasets, including IrishQA, our curated question-answering dataset on topics surrounding Ireland and its cultural nuances, available in both English and Irish; and MT-bench, translated to Irish and verified by native Irish speakers.
    \end{itemize}

\vspace{-0.5cm}
\section{\ourmodel{}: Irish-eXcellence Large Language Model}
    \emph{Data Collection and Pre-processing.} For pre-training, we meticulously curate all available open-source Irish data. Our primary sources include CulturaX \cite{nguyen2023culturax} and Glot500 \cite{imanigooghari-etal-2023-glot500}, both of which provide valuable content from multilingual websites, including a subset dedicated to Irish. Additionally, we incorporate data from the Irish segment of the ga-en bitext pair of ParaCrawl v7 \cite{banon-etal-2020-paracrawl}, text sourced from Irish Wikipedia, and data from Corpora Irish \cite{goldhahn-etal-2012-building}. To ensure data integrity, we conduct thorough cleaning and filtering, following the heuristics introduced in~\cite{luukkonen-etal-2023-fingpt}. This is then followed by the removal of duplicates across datasets using n-gram ($n=5$ in this work) matching, as there are potential overlaps between data sources. This pre-processing process ensures a high quality corpora of data, particularly important in resource-constrained environments \cite{Raffel2019ExploringTL}. The size and proportion of each dataset in the final data mixture are summarized in Table~\ref{tab:pretraindata}. We believe our mixture is relatively larger than all previous works for Irish NLP~\cite{mille-etal-2023-generating,gaBert}.

    \begin{table}[!ht]
    \centering
    \caption{Pre-train data statistics.}
    \begin{tabular}{@{}lP{2cm}P{2cm}r@{}}
    \toprule
    \textbf{Dataset}             & \textbf{Chars (before pre-processing)} & \textbf{Chars (after pre-processing)} & \textbf{Ratio} \\ \midrule
    CulturaX-ga \cite{nguyen2023culturax} & 1.3B & 1.2B & 65.0\% \\
    Glot500-ga \cite{imanigooghari-etal-2023-glot500}  & 1.3B & 483.7M & 27.1\% \\
    Gaparacrawl \cite{banon-etal-2020-paracrawl} & 395.2M & 106.3M & 6.0\% \\
    Gawiki\protect\footnotemark & 41.9M & 23.4M & 1.3\% \\
    Corpora Irish \cite{goldhahn-etal-2012-building} & 37.1M & 11.1M & 0.6\%
    \\ \midrule
    Total & 3.1B & 1.8B & 100\%
    \\ \bottomrule
    \end{tabular}
    \label{tab:pretraindata}
    \end{table}

    \footnotetext   {We use the dumps from \url{https://dumps.wikimedia.org/gawiki/20240201/}}

    \emph{Continued Pre-training.} Given the constraints of extremely low-resource languages, where the amount of mono-Irish data is insufficient for the pre-training of LLMs from scratch, we adopt a strategy of language adaptation for continued pre-training. Starting from a pre-trained LLM, we iteratively enhance its proficiency in understanding additional languages by exposing it to target language data. We follow the hypothesis from recent works~\cite{yong-etal-2023-bloom,xu2024a} that we can transfer world knowledge and English proficiency to Irish through this process. Moreover, we employ a tailored data scheduler to feed parallel English-Irish data first, facilitating the establishment of linguistic connections between the two languages. Parallel pairs of English-Irish sentences from ELRC~\cite{european2020elrc} are leveraged. The parallel data is approximately 1\% of the size of the mono-Irish corpus introduced in Table~\ref{tab:pretraindata}. Then, the larger set of mono-Irish is fed to the model to further refine the proficiency of the language and coverage of Irish cultural nuances.

    \emph{Model Architecture.} The selection of an appropriate base model is vital for maximizing support for the Irish language. We employ perplexity as the main metrics to compare between base models. We select a subset of the pre-training corpora with $\approx2M$ characters, named \pretrainsubset{}, to compute the metric on. We identify Llama 2-13B as the optimal base model. With a lower perplexity score (8.94) compared to alternative models such as Mistral-7B~\cite{abs-2310-06825}(11.68) or BLOOM-7B~\cite{bloom}(63.75), Llama 2-13B serves as a strong foundation. Additionally, we only consider models of adequate size (less than 20B parameters), as we fear the problem of underfitting larger models.
    
    \emph{Tokenizer's Vocabulary Expansion.} We extend the original vocabulary of the chosen LLM, as it contains mostly tokens for languages such as English. From the original size of 32k tokens, the vocabulary is expanded to include 10k Irish tokens. This not only allows for efficiency improvement in generating Irish text but also maintains the same generation speed for English. Moreover, the expanded vocabulary provides a more meaningful representation of Irish text and enables bilingual capabilities in the final LLM.

    \emph{Supervised Instruction Fine-tuning.} After pre-training phase, to align the model more closely with human preferences, we apply Supervised Instruction Fine-tuning~\cite{mishra-etal-2022-cross,chatgpt}. This process enhances the model's ability to follow human instructions effectively. We denote the pre-trained model as \ourmodel{}, and the instruction fine-tuned version as \ourmodel{}-Instruct.

    \emph{Evaluation Data.} In evaluating pre-trained LLMs, our focus lies on assessing their language understanding capabilities. We utilize various existing NLP benchmarking datasets tailored for Irish, including the Cloze Test \cite{gaBert}, the Irish subset of SIB-200 \cite{adelani-etal-2024-sib}, and gaHealth \cite{lankford-etal-2021-machine} However, we observe a lack of natural language generation benchmarking tasks, such as question-answering, and Irish-specific cultural content in these datasets. To address this gap, we introduce \emph{IrishQA}, a question-answering dataset on topics surrounding Ireland and its cultural nuances. The dataset has a total of 60 question-answer pairs, each pair is accompanied by a paragraph, containing the contextual information necessary to answer the question that the LLMs will be prompted with alongside the question. The dataset is available in both English and Irish, created by native Irish speakers. Furthermore, for the evaluation of instruction-tuned LLMs, we follow a recent approach \cite{zheng2023judging} that highlights the potential of using really large language models, such as GPT-4, as an automated judge. We translate and adapt the MT-bench dataset to Irish using Google Cloud Translation API~\cite{googleCloudTranslation}, carefully verifying and correcting the translation through native Irish speakers.

\vspace{-0.2cm}
\section{Benchmarking Results}
    \begin{table*}[!ht]
    \centering
    \caption{Evaluation results of pre-trained models on curated set of Irish benchmarking datasets. \ourmodel{}$_{npd}$ is our approach but without training the base model on parallel data first, and \ourmodel{}$_{nte}$ is the ablation study without tokenizer's vocabulary expansion.}
    \begin{tabular}{@{}p{2.2cm}P{1.8cm}P{1.8cm}P{1.8cm}P{1.8cm}P{1.8cm}P{1.8cm}P{1.8cm}P{1.8cm}@{}}
    \toprule
    \textbf{Model} & \specialcelltwo{\textbf{Acc. on Cloze Test}\\\textbf{(0-shot)}} & \specialcelltwo{\textbf{Acc. on SIB-200 (Irish subset)}\\\textbf{(10-shot)}} & \specialcelltwo{\textbf{Exact-match on IrishQA (ga)}\\\textbf{(5-shot)}} & \specialcelltwo{\textbf{BLEU-4 on gaHealth - en2ga}\\\textbf{(5-shot)}} & \specialcelltwo{\textbf{BLEU-4 on gaHealth - ga2en}\\\textbf{(5-shot)}} & \textbf{Tokenizer speed  on \pretrainsubset{} (chars/token)} & \textbf{Fertility rate on \pretrainsubset{} (tokens/word)} \\ \midrule
    gpt-3.5-turbo                                                                                                   & N/A                             & N/A                                                         & 0.2222                                                 & {0.1864}                                 & {0.4257} & 2.45 & N/A                                 \\
    Llama 2-70B & 0.63 & 0.7059 & 0.2963 & 0.0852 & 0.3861 & 2.26 & 1.95 \\
    Llama 2-13B                                                                                       & 0.54                            & 0.5343                                                      & {0.3148}                                           & 0.0325                                       & 0.2560  & 2.26 & 1.95                                      \\
    BLOOM-7B1 & 0.45 & 0.1471 & 0.0000 & 0.0061 & 0.0184 & 2.62 & 1.80 \\
    \midrule
    \ourmodel{}$_{npd}$ (Ours) & 0.61 & 0.7206 & {\ul 0.3704} & 0.2950 & 0.4563  & 3.47 & 1.57\\
    \ourmodel{}$_{nte}$ (Ours) & \textbf{0.80} & {\ul 0.7402} & 0.3333 & {\ul 0.3297} & {\ul 0.4631} & 2.26 & 1.95 \\
    \ourmodel{} (Ours) & {\ul 0.75}                      & \textbf{0.7794} & \textbf{0.3889}                                        & \textbf{0.3334}                              & \textbf{0.4636} & 3.47 & 1.57                              \\ \bottomrule
    \end{tabular}
    \label{tab:irishexp}
    \end{table*}

    \emph{Irish Performance.} The evaluation results in Table~\ref{tab:irishexp} reveals the robust performances of \ourmodel{}, surparssing larger models across datasets, with the largest gap of 12\% accuracy on Cloze Test, compared to the second best model. These margins remain wider when compared against baselines of similar sizes (Llama 2-13B and BLOOM-7B1). Notably, baseline models seem to have an understanding of the Irish language, as evidenced by their ability to translate from Irish to English (gaHealth-ga2en task), but they failed to generate coherent Irish text, resulting in superior performance on the gaHealth-ga2en task compared to gaHealth-en2ga. Nevertheless, our model enhances the performance across both translation directions, with substantial improvements of 0.1471 BLEU-4 on gaHealth-ga2en and 0.0379 BLEU-4 on gaHealth-en2ga. Furthermore, The results on IrishQA and the examples from Figure~\ref{fig:example} show that the baseline models are proned toward inherent biases, as their responses while in Irish, usually unrelated to the posed questions. We note that some results for the closed-source gpt-3.5-turbo model are not available, as they do not provide loglikelihood information in their APIs, which is required for some benchmarking tasks~\cite{eval-harness}.
    
    \emph{Catastrophic Forgetting.} While our models exhibit robust performance in Irish-related tasks, we notice instances of catastrophic forgetting when benchmarked on English-related datasets, as demonstrated in Table~\ref{tab:englishexp}. This might be due to the iterative training on Irish corpora, leading the model to forget the knowledge as previously trained on English dataset. This highlights a potential area for further improvement in model adaptation and fine-tuning techniques.

    \begin{table}[!t]
    \centering
    \caption{Ablation Experiment - Evaluation results of pre-trained models on English benchmarking datasets.}
    \begin{tabular}{@{}lP{1.2cm}P{1.2cm}P{1.2cm}P{1.2cm}@{}}
    \toprule
    \textbf{Model} &
      \specialcell{\textbf{Exact-match on Natural Question}\\\textbf{(5-shot)}} &
      \specialcell{\textbf{Exact-match on IrishQA (en)}\\\textbf{(5-shot)}} &
      \specialcell{\textbf{Acc. on Winogrande}\\\textbf{(5-shot)}} &
      \specialcell{\textbf{Acc. norm on HellaSwag}\\\textbf{(10-shot)}} \\ \midrule
    gpt-3.5-turbo              & \textbf{0.4660} & 0.3333                & N/A                   & N/A             \\
    Llama-2-70B & {\ul 0.3806} & {\ul 0.4074} & \textbf{0.8374} & \textbf{0.8701} \\
    Llama 2-13B  & {0.3069} & \textbf{0.4444}       & {\ul 0.7609}          & {\ul 0.8223} \\
    BLOOM-7B & 0.0806 & 0.1667 & 0.6519 & 0.6202 \\
    \midrule
    \ourmodel{}$_{npd}$ (Ours) & 0.1848 & 0.3704 & 0.7017 & 0.7695 \\
    \ourmodel{}$_{nte}$ (Ours) & 0.1584 & 0.3704 & 0.6969 & 0.7690 \\
    \ourmodel{} (Ours)         & 0.1668 & {0.3704} & 0.7135       & 0.7758       \\ \bottomrule
    \end{tabular}
    \label{tab:englishexp}
    \end{table}
    
    \begin{table}[!t]
    \centering
    \caption{Ablation Experiment - Evaluation of instruction-tuned models on English and Irish versions of MT-bench.}
    \begin{tabular}{@{}lcc@{}}
    \toprule
    \textbf{Model}             & \textbf{MT-Bench (en)} & \textbf{MT-Bench (ga)} \\ \midrule
    gpt-3.5-turbo              & \textbf{8.63} & \textbf{5.47}                       \\
    Llama 2-70B-chat  & {\ul 6.86} & 2.22 \\
    Llama 2-13B-chat  & 6.65 & 1.94                        \\
    \midrule
    \ourmodel{}-Instruct (Ours)         & 5.82 & {\ul 5.01}                        \\ \bottomrule
    \end{tabular}
    \label{tab:mtbench}
    \end{table}

    \emph{Instruction Following Capability.} We follow the approach outlined in~\cite{zheng2023judging} and introduce our Irish-translated version of MT-bench, which is a challenging multi-turn question set spanning 8 categories. We evaluate the instruction-following capability of \ourmodel{}-Instruct on both the original English and our proposed Irish versions. As illustrated in Table~\ref{tab:mtbench}, \ourmodel{}-Instruct demonstrates sophisticated performance on the Irish version, nearly rivaling that of gpt-3.5-turbo (5.47 compared to 5.01), despite being more than 10 times smaller in size. However, we also notice a performance degradation on English version, compared to the Llama 2-13B-chat, which is of the same size as ours. This aligns with the observation of catastrophic forgetting.

    \emph{Generation Efficiency.} To assess the efficiency gained through the expanded vocabulary with native Irish tokens, we compute two metrics: tokenizer speed and fertility rate. Tokenizer speed, representing the proportion of split tokens to the number of characters (chars/token), and fertility rate, indicating the average number of tokens required to represent a word or document (tokens/word), are computed within the vocabulary as the proportion of the number of tokens to the tokens that start with "space" character. As reported in Table~\ref{tab:irishexp}, our expanded tokenizer exhibits a generation speed 1.54 times faster (3.47) than the base Llama 2's tokenizer (2.26), establishing it as the fastest among models. Moreover, as seen in comparisons between \ourmodel{} and \ourmodel{}$_{nte}$ in Table~\ref{tab:irishexp} and \ref{tab:englishexp}, the expansion of vocabulary also leads to reliable performance enhancement.

    \emph{Ablation Study Without Parallel Data.} We carry out a training experiment with the same base LLama 2-13B LLM, but without the scheduling of parallel data at the start of continued pre-training, denoted as \ourmodel{}$_{npd}$. The results in Table~\ref{tab:irishexp} and~\ref{tab:englishexp} highlight the usefulness of our proposed approach of utilizing parallel corpora to bridge the gaps of linguistic differences between the two languages.

\section{Conclusion}
    In this paper, we develop a novel Irish-based LLM, \ourmodel{}, to challenge the task of training LLMs for low-resource languages, particularly Irish. Through meticulous data curation and cleaning and innovative training strategies and techniques, we have demonstrated the feasibility of creating robust LLMs despite constrained resources. Our model showcases impressive proficiency in understanding and generating Irish text, with notable achievements in various benchmarking tasks. We contribute comprehensive benchmarking datasets, including IrishQA and Irish version of MT-bench, which enable thorough evaluation and facilitate future research. We believe that our work has important implications for preserving and promoting the Irish language, and can serve as a blueprint for adapting LLMs to other low-resource languages, helping to bridge the gap between high- and low-resource languages in NLP.

\ack We would like to acknowledge CloudCIX Limited for the generous collaborative support of computing resources on their NVIDIA HGX/H100 GPU cluster. This research work has emanated from research conducted with financial support from Science Foundation Ireland under Grant 12/RC/2289-P2 and 18/CRT/6223. 

\newpage

\bibliography{ecai}
\end{document}